\documentclass{article}

\usepackage[preprint]{neurips_2026}

\usepackage[utf8]{inputenc} 
\usepackage[T1]{fontenc}    
\usepackage{hyperref}       
\usepackage{url}            
\usepackage{booktabs}       
\usepackage{amsfonts}       
\usepackage{nicefrac}       
\usepackage{microtype}      
\usepackage{xcolor}         

\usepackage{amsmath,amsfonts,bm}

\def\eqref#1{equation~\ref{#1}}

\def\1{\bm{1}}

\DeclareMathAlphabet{\mathsfit}{\encodingdefault}{\sfdefault}{m}{sl}
\SetMathAlphabet{\mathsfit}{bold}{\encodingdefault}{\sfdefault}{bx}{n}

\usepackage{bbm}
\usepackage{wrapfig}
\usepackage{adjustbox}
\usepackage{multirow, makecell, booktabs}
\usepackage{siunitx}
\usepackage{graphicx}
\usepackage{caption}
\setcellgapes{3pt}
\makegapedcells
\usepackage{spverbatim}
\usepackage{fancyvrb}
\usepackage{listings}
\newcommand{\eat}[1]{} 

\usepackage{microtype}
\usepackage{graphicx}
\usepackage{subcaption}
\usepackage{booktabs} 

\usepackage{amsmath}
\usepackage{amssymb}
\usepackage{mathtools}
\usepackage{amsthm}

\usepackage[capitalize,noabbrev]{cleveref}

\usepackage{hyperref}
\usepackage{algorithm}
\usepackage{algorithmic}
\usepackage{graphicx}

\usepackage{tcolorbox}
\tcbuselibrary{breakable,skins}
\usepackage{xcolor}
\usepackage{booktabs}
\usepackage{array}

\title{Subagents vs Agent Skills: Executing Reusable Knowledge for Long-Horizon Agentic Tasks}

\author{
  \textbf{Wasu Top Piriyakulkij}\textsuperscript{*}$^1$ \qquad 
  \textbf{Rachel Lawrence}$^2$ \qquad 
  \textbf{Alicia Curth}$^2$ \\ 
  \textbf{Sushrut Karmalkar}$^2$\qquad
  \textbf{Niranjani Prasad}$^2$\\
  Cornell University$^1$ \quad Microsoft Research Cambridge$^2$
}

\begin{document}

\maketitle

\begin{abstract}
How can language model agents effectively leverage libraries of reusable knowledge to solve long-horizon tasks? 
Recent work has increasingly focused on agent skills: reusable capabilities represented as skill packages, i.e., multi-file bundles containing instructions, scripts, and other resources that help agents perform specific tasks.
Agent skills are typically executed by loading their skill instructions into an agent's context and relying on the agent to follow them.
As task horizons grow, however, this approach becomes increasingly brittle, because reasoning quality degrades as more information accumulates in the context window. 
We investigate an alternative approach in which skill packages are instead invoked as subagents. Rather than loading skill instructions into the main context, subagent execution spawns fresh context windows dedicated to solving individual subtasks. 
We show that subagent execution outperforms agent-skill execution when skill packages expose clear input-output contracts and their instructions encode the procedural knowledge needed to fulfill those contracts.
The tradeoff is additional communication overhead, as extra tokens are required to coordinate between the main agent and its subagents.
Our results show that the benefit of reusable knowledge depends not only on its content, but also on how it is organized and invoked.
\end{abstract}

\renewcommand{\thefootnote}{\fnsymbol{footnote}}
\footnotetext[1]{Work done during an internship at Microsoft Research Cambridge.}
\renewcommand{\thefootnote}{\arabic{footnote}}

\section{Introduction}

How do we equip AI agents with domain-specific knowledge so that they can solve complex tasks? 
This is a longstanding question in AI, dating back to the symbolic AI era, where researchers attempted to encode knowledge explicitly through structured representations such as rules, logic, and knowledge graphs \cite{shortliffe1976mycin, lenat1995cyc, brachman2004knowledge, speer2017conceptnet}. 
The rise of deep learning shifted the paradigm from direct knowledge injection to weight-based learning; knowledge is acquired from data and stored implicitly in neural weights.
The advent of large language models (LLMs) has partially shifted the field back toward explicit knowledge injection. 
Through in-context learning, LLMs acquire and apply new information at inference time, without weight updates, in one of the most expressive symbolic representations: natural language.

Recently, \emph{agent skills} \cite{anthropic2025skills} have emerged as an effective way of knowledge injection for LLM agents. 
Each agent skill is represented by a \emph{skill package}: a multi-file package containing instructions and scripts that are useful for solving specific tasks. 
The names and descriptions of available skills are then given to the LLM agent. 
When the agent invokes a skill, it gains access to that skill's instructions and follows them.
These agent skills have become the dominant way of providing LLM agents with reusable, domain-specific knowledge.

Despite their name, agent skills are quite different from the notion of a skill in reinforcement learning \cite{sutton1999between, parr1997reinforcement, dietterich2000hierarchical}: a skill is a temporally extended policy invoked by a higher-level controller. 
Invoking agent skills, instead, simply loads the skill recipes (SKILL.md files) into the LLM agent's context, rather than directly performing the skill or solving a subtask. 
Thus, even when an agent skill contains relevant information, invoking it might be unproductive, as adding more instructions increases context length, and agent performance degrades as context grows \cite{liu2024lost, du2025context, hong2025exploring, li2026tool}. 
This issue of bloated context becomes increasingly problematic as task horizons become longer.

\begin{figure*}[t]
\centering
\includegraphics[width=0.85\linewidth]{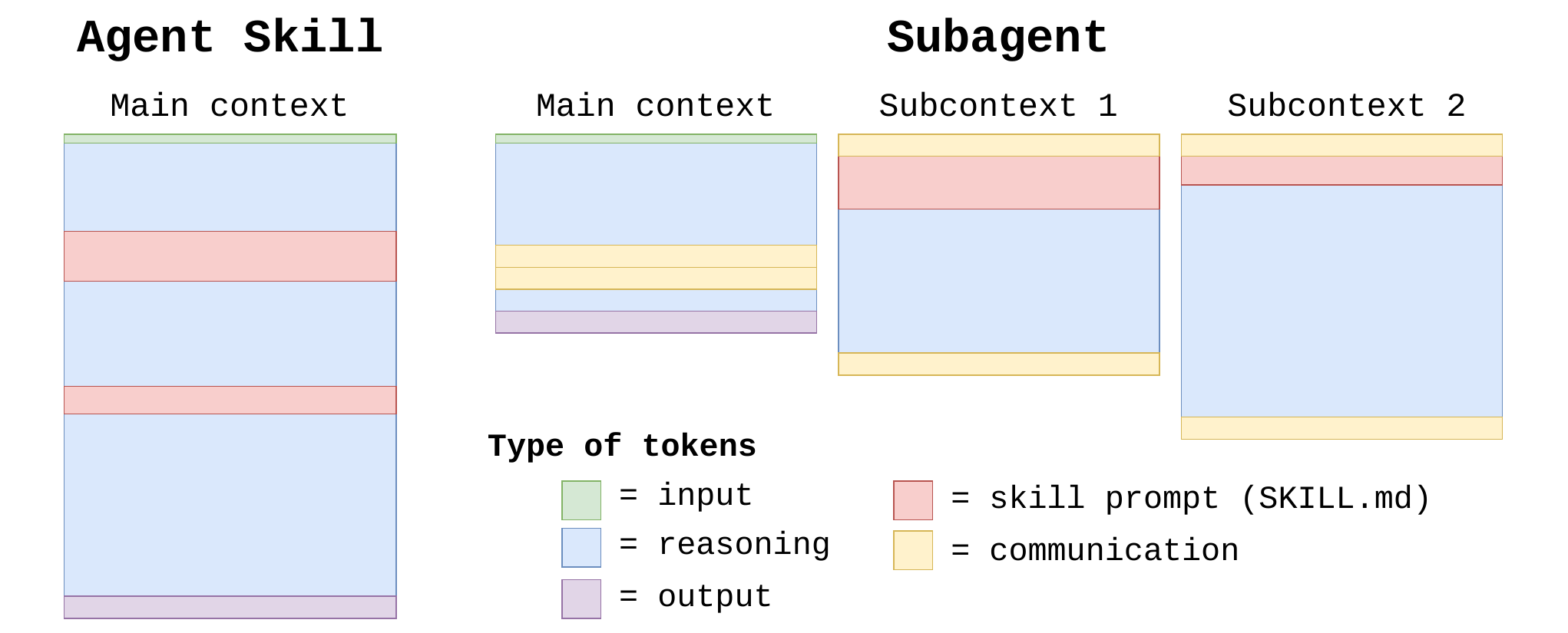}
\caption{Two type of skill package execution: agent skill vs subagent. 
While agent-skill execution does all the reasoning in a single context, subagent execution creates new context windows for each subtasks, each window initialized with skill instructions (SKILL.md).
Since subagents reason within their own context, however, extra tokens are needed for the main agent and subagents to communicate.}
\label{fig:overview}
\end{figure*}

We therefore investigate an alternative approach to agent skills: executing skill packages as \textit{subagents}. 
Instead of loading skill instructions into the main context, subagent execution spawns a new context window seeded with the skill instructions, performs the skill independently, and returns only the output to the main agent.
We possess that by dividing a task across multiple context windows, the maximum amount of information processed by any individual context can be reduced (\Cref{fig:overview}).
While existing agent harnesses do have subagents, they are generally underutilized, targeting mainly situations where independent subtasks can be run in parallel to reduce latency \cite{anthropic2026subagents}.
Executing skills as subagents has downsides, however. 
For instance, we lose the ability to combine knowledge from multiple skills to solve a subtask.


In this work, we show that a skill package is suitable for subagent execution when it is presented as procedural knowledge with clear input and output contracts, analogous to the options framework in reinforcement learning\cite{sutton1999between}.
Procedural skills with well-defined input-output interfaces are inherently self-contained, allowing their internal reasoning to be delegated to separate contexts.
On SkillsBench, a benchmark for evaluating agent skills, we synthesize procedural, contract-driven skill packages derived from successful task trajectories and show that executing these skill packages as subagents significantly outperforms agent-skill execution.
Our findings suggest that both how skills are executed and how skill knowledge is organized are important design choices that substantially impact agent performance.

\section{Background: Agents with Tools, Agent Skills, and Subagents}

We are interested in a tool-calling language model agent that operates iteratively. 
Let $c_t$ denote the agent's context at turn $t$. For notational simplicity, we assume the agent invokes one tool per turn. At each turn $t$, the language model $\pi_{LLM}$ generates a text response $r_t$, a tool $k_t$, and tool arguments $x_t$:
\begin{equation}
    (r_t, k_t, x_t)=\pi_{LLM}(c_t)\label{eq:1}
\end{equation}
The selected tool is executed by a tool executor $E$:
\begin{equation}
    o_t = E(k_t, x_t)\label{eq:2}
\end{equation}
where $o_t$ denotes the tool output. 
The agent context is then updated by appending the response, tool call, and tool result: 
\begin{equation}
    c_{t+1} = c_t \oplus (r_t, k_t, x_t, o_t)\label{eq:3}
\end{equation}
We slightly abuse notation and use $\oplus$ to denote appending an interaction trace to the context. Any structured objects are assumed to be converted into their textual representations before concatenation.
The agent repeats this process until a final answer is produced at some turn $T$.

\paragraph{Agent Skills.} 
An agent skill is specified by a skill package $s$:
\begin{equation} 
    s = (d,m,R)\label{eq:4}
\end{equation}
where the description $d$ provides a concise summary of the skill's purpose and is always given to the main agent for skill discovery, the instruction file $m$ (implemented as \texttt{SKILL.md}) specifies how the skill can be performed, and the resource collection $R$ contains supporting artifacts such as scripts, examples, reference documents, and nested directories.

An agent skill is a tool constructed from a skill package, $k = \mathrm{AgentSkill}(s)$, and we write $m_k$ for the instruction file of its underlying package.
Invoking an agent skill requires no additional arguments and simply returns the skill instructions $m_k$: 
\begin{equation} 
E(k=\mathrm{AgentSkill}(s),\varnothing) = m_k.\label{eq:5}
\end{equation}
The skill contents are thus exposed directly to the main context, and the procedure prescribed by $m_k$ is carried out by $\pi_{LLM}$ itself: 
each step of the skill's execution is an ordinary agent step in the main context $c$, applying \Cref{eq:1,eq:2,eq:3}

\paragraph{Subagents.} 
Subagents provide an alternative mechanism for utilizing reusable knowledge. 
Like agent skills, a subagent is constructed from a skill package, $k = \mathrm{Subagent}(s)$.
Unlike agent skills, however, invoking a subagent does not expose the skill instruction file directly to the main agent, nor is its execution carried out by $\pi_{LLM}$.
Instead, the tool executor instantiates a distinct policy that conditions on the skill instructions $m_k$ and initializes a new agent context from the task input $x_t$:
\begin{equation}
    \pi^{(k)}(\cdot) \;\triangleq\; \pi_{LLM}( m_k \oplus\, \,\cdot\,),
    \qquad
    c^{(k)}_0 = x_t. \label{eq:6}
\end{equation}
The subagent then executes an independent tool-calling reasoning process within this separate context, following \Cref{eq:1,eq:2,eq:3} under $\pi^{(k)}$, and returns only the response $r^{(k)}_T$ from its last turn $T$: 
\begin{equation} E(k=\mathrm{Subagent}(s),x_t) = r^{(k)}_T. \label{eq:8}\end{equation}

Both mechanisms draw on the same skill package $s$, but they differ in which policy that knowledge conditions.
Agent skills let a single policy $\pi_{LLM}$ operating over a single context, with the skill's execution unfolding as part of the main agent's own turn sequence; subagents introduce a separate subpolicy $\pi^{(k)}$ per skill package, each acting on a context the main agent neither reads nor writes.
Note that $\pi^{(k)}$ is defined here over the same LLM as the main $\pi_{LLM}$, but it can instead be instantiated from a different one; what makes it a separate policy is that it is separately instantiated and communicates with the main agent only through the input and output, $x_t$ and $r^{(k)}_T$.

\section{Subagents For Long-horizon Agentic Tasks}\label{sec:subagents_method}

LLM reasoning capability declines with context length \cite{liu2024lost, du2025context, hong2025exploring, li2026tool}; this decline has been linked to bandwidth-limited attention: LLMs can only communicate a bounded amount of information across long inputs \cite{schnabel2026lost}.
Long-horizon agentic tasks are particularly susceptible to this degradation: they require agents to reason over increasingly long trajectories, consisting of verbose tool inputs and outputs, and intermediate reasoning.
While skill packages may contain useful knowledge to solve such tasks, invoking an agent skill, which loads the skill instructions into the agent's context, amplifies the context overload issue.
Solving long-horizon tasks may depend not only on having access to useful skills, but also on executing those skills in a way that controls context growth.
To achieve this, we explore the use of 
\textit{subagents}.

\subsection{Reducing Peak Context Length Through Subagent Execution}\label{sec:reduce_peak}

To control context growth, we directly ask whether spawning more context windows can help. What matters for bandwidth limitations discussed above is not total context consumed across a task, but the \emph{peak} context length any single window must process. 
If we can decompose a task into smaller, self-contained subtasks and execute each subtask within a separate context window, we expect the peak context length across all these windows to be shorter than that of a single, monolithic context window. 
Subagent execution encodes this exact mechanism (\Cref{fig:overview}). 
When a subagent is invoked, a new context window is spawned and initialized with the subagent's input and its skill package's instructions.


One way to view subagent execution is through the lens of information encapsulation. 
Subagents encapsulate subtask-specific information.
Subagents' internal trajectories are hidden away from the main, parent agent. 
Only the final output of the subagent is visible to the main agent.
The main agent then has less information to reason over, and so it can reason better.

However, peak context reduction with subagents comes at a cost. 
Executing a task across multiple context windows introduces communication overhead between the main agent and subagents (\Cref{fig:overview}): since subagents do not see the main context, they must be given sufficient input information to carry out their subtask, which means relevant information is often repeated across context windows.
Subagent execution thus trades a higher total token count for reduced peak context length.

We note that popular agent harnesses such as Claude Code and OpenAI Codex do support spawning subagents, but they tend to be used for task parallelization rather than encapsulating information \cite{anthropic2026subagents}. 
In many cases, the spawned subagents do not use any skill package; they function as lightweight parallel workers. 
The use of subagents as a mechanism for executing reusable knowledge remains relatively underexplored.

\subsection{Input-Output Contracts for Effective Subagents}\label{sec:criteria}

Reducing peak context length with subagents is useful for agent performance \emph{only if} the subtasks in the subcontext windows are actually solved correctly. 
Unlike agent skills, subagents introduce an additional subtask delegation problem: the main agent must identify an appropriate subagent for a subtask and provide sufficient information for the subagent to solve it successfully.
If the main agent fails at either of these, subagents become ineffective.

To ensure the main agent correctly delegates subtasks and provides required information for each subtask, each skill description needs to tell the main agent what subtask the corresponding subagent can solve and what information should be supplied to it. 
Of course, the skill instructions then must match the skill description: they have to tell the subagents how to solve the subtasks the description promises to solve. 

Skill packages designed this way echo the options framework \cite{sutton1999between}: invoked as subagents, they behave as language-based options. 
An option is defined as \begin{equation} \omega = (\mathcal I_\omega,\pi_\omega,\beta_\omega) \label{eq:option_format} \end{equation} where \(\mathcal I_\omega\) is the initiation set, \(\pi_\omega\) is the option policy, and \(\beta_\omega\) is the termination condition. An option may be invoked only in states belonging to \(\mathcal I_\omega\), after which it follows \(\pi_\omega\) until termination occurs according to \(\beta_\omega\). 

Similarly, an effective subagent skill package should specify when the subagent can be appropriately invoked, how it solves the delegated task, and what information it returns upon completion. We therefore structure the skill description as 
\begin{equation} 
d = (q_{\mathrm{in}}, h, q_{\mathrm{out}}) \label{eq:subagent_format} 
\end{equation} 
where \(h\) is a concise summary of the skill's purpose, and \(q_{\mathrm{in}}\) and \(q_{\mathrm{out}}\) are natural-language input and output contract specifications, respectively. 

Substituting into \Cref{eq:4}, the skill package is
\begin{equation}
s = \bigl((q_{\mathrm{in}}, h, q_{\mathrm{out}}),\, m,\, R\bigr)
\label{eq:subagent_package}
\end{equation}
Just as an option implements \(\omega : \mathcal I_\omega \rightarrow \mathcal
T_\omega\), where \(\mathcal T_\omega\) is the set of terminal states permitted
by \(\beta_\omega\), the contracts define sets of valid inputs \(\mathcal
X_{\mathrm{in}}\) and outputs \(\mathcal X_{\mathrm{out}}\), and the resulting
subagent implements \(\mathrm{Subagent}(s) : \mathcal
X_{\mathrm{in}} \rightarrow \mathcal X_{\mathrm{out}}\). The input contract
\(q_{\mathrm{in}}\) plays the role of the initiation set \(\mathcal I_\omega\),
specifying the conditions under which the subagent may be invoked. 
The instruction file \(m\) plays the role of the option policy
\(\pi_\omega\), encoding the procedure that carries a valid input to an output.
\(q_{\mathrm{out}}\) plays the role of the termination condition
\(\beta_\omega\), constraining what the subagent returns to the main agent.
These contracts define a self-contained interface through which information
enters and leaves the subagent.
We therefore hypothesize that procedural instructions coupled with explicit input-output contracts enable effective subagent execution.

\section{Experiments}\label{sec:exp}

We now test the claims made in \Cref{sec:subagents_method}. 
\Cref{sec:criteria} claims that subagents help only when skill packages expose explicit input-output contracts and pair them
with instructions that faithfully deliver on those contracts, and \Cref{sec:reduce_peak}
claims that subagents then trade higher total tokens for lower peak context. 
We test the contract claim first, comparing both execution strategies on existing curated and our contract-based skill packages.
We then increase context pressure by adding distracting tools, testing whether the advantage of subagents grows with context length. 
Finally, we measure peak context and total token cost directly.

\paragraph{Domain: SkillsBench.} SkillsBench \cite{li2026skillsbench} is a benchmark specifically designed to evaluate the effectiveness of agent skills, comprising 87 long-horizon agentic tasks across diverse domains. 
Each task comes with a set of human-authored skill packages, which lets us test agent-skill and subagent execution under a curated set of packages.

We use OpenHands \cite{wang2025openhands} as the agent harness throughout this paper, with limited modifications to the original OpenHands SDK; details can be founded at \Cref{app:openhands}.

\paragraph{Synthesizing procedural skill packages with explicit input-output contracts.}
From inspecting the curated SkillsBench skill packages, we find that very few satisfy the criteria for effective subagent execution discussed in \Cref{sec:criteria}. In particular, most packages describe relevant information but do not clearly specify the inputs expected by the skill or the outputs it should produce.

To construct skill packages better suited for subagent execution, we create a new set of procedural skill packages with explicit input-output contracts for each task. 
Specifically, we first run OpenHands agents \cite{wang2025openhands} using GPT-5.3 Codex on all SkillsBench tasks for three independent runs, collect successful trajectories, and use Copilot CLI \cite{githubcopilotcli}, with minimal human intervention, to synthesize skill packages. 
Using this procedure, we successfully synthesize these well-interfaced procedural skill packages for 64 of the 87 benchmark tasks.
More details can be founded at \Cref{app:skill_learning}.
The results of this paper are on this 64-task subset of SkillsBench.
From now on, we refer to this skill package set as the \emph{synthesized} set.

\begin{figure*}[t]
\centering
\includegraphics[width=.9\linewidth]{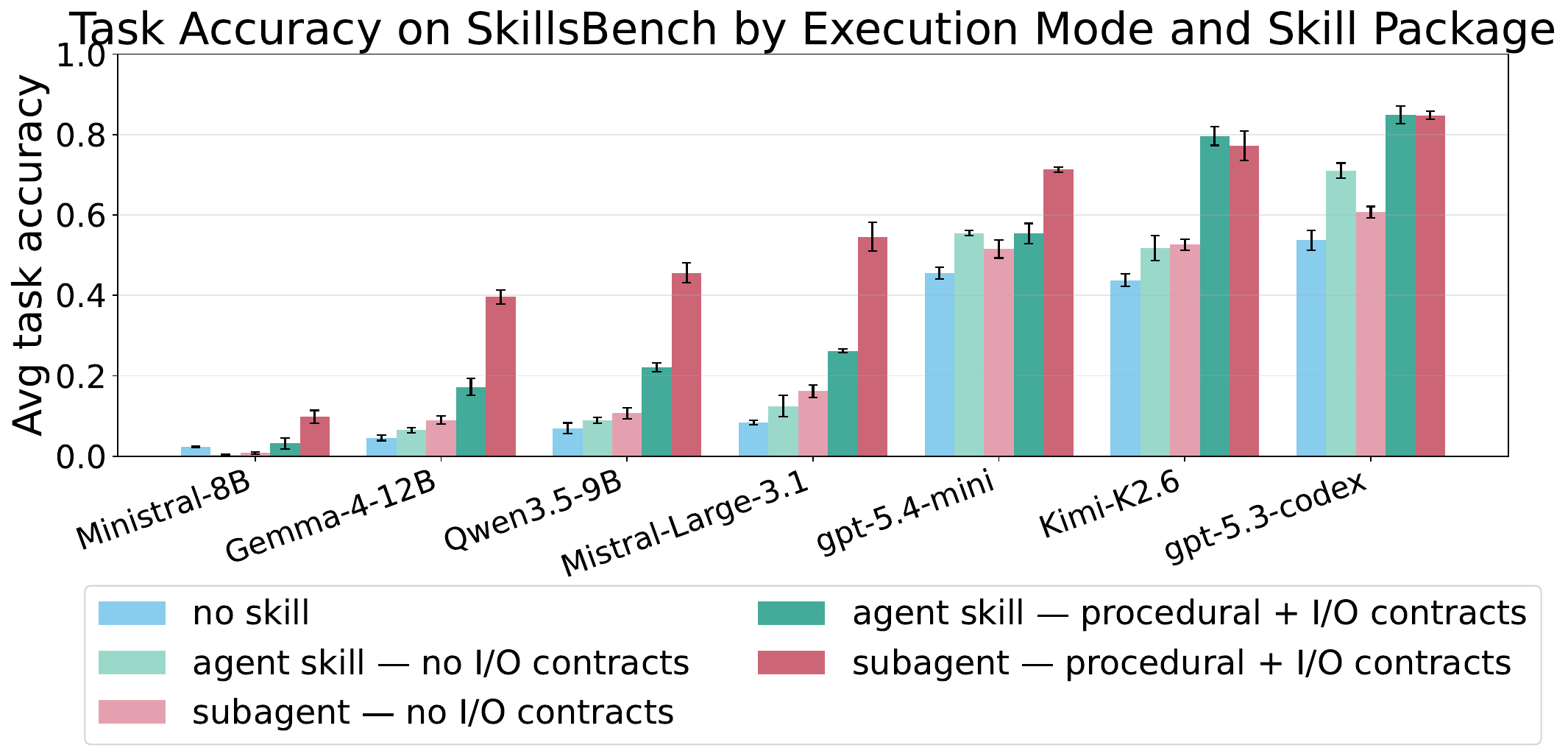}
\caption{Average task accuracy on SkillsBench for each execution mode (agent skill /
subagent) and skill package set, across base LLM models. The \emph{no I/O contracts} packages are
the human-authored ones curated with SkillsBench, which describe relevant knowledge but rarely
specify a skill's expected inputs or outputs. The \emph{procedural + I/O contracts} packages
are ours, synthesized from successful trajectories to give procedural instructions with
explicit input-output contracts.}
\label{fig:main}
\end{figure*}

\paragraph{Main results.}
\Cref{fig:main} shows the main comparison between agent skills and subagents. 
When using the original curated SkillsBench skills, which lack input-output contracts, agent skills match or outperform subagents across all models.
However, the trend reverses when using our procedural skill packages with explicit input-output contracts. 
On these skill packages, subagents outperform agent skills, with the gain being largest for smaller models that are more bandwidth limited.

This result supports our central hypothesis: subagent execution becomes beneficial when skills are designed as procedural abstractions with well-defined interfaces. 
In contrast, when skills primarily contain loosely structured knowledge, directly injecting the skill contents into the main agent's context, i.e., agent skill execution, is more effective.

Finally, our synthesized skill packages outperform the curated ones across both execution modes. 
This is not a controlled comparison, since the two sets differ in content, but it suggests that the synthesis procedure produces skills of at least comparable quality to the curated ones.

\begin{wrapfigure}[19]{r}{0.55\linewidth} 
\centering
\vspace{-1em}
\includegraphics[width=0.95\linewidth]{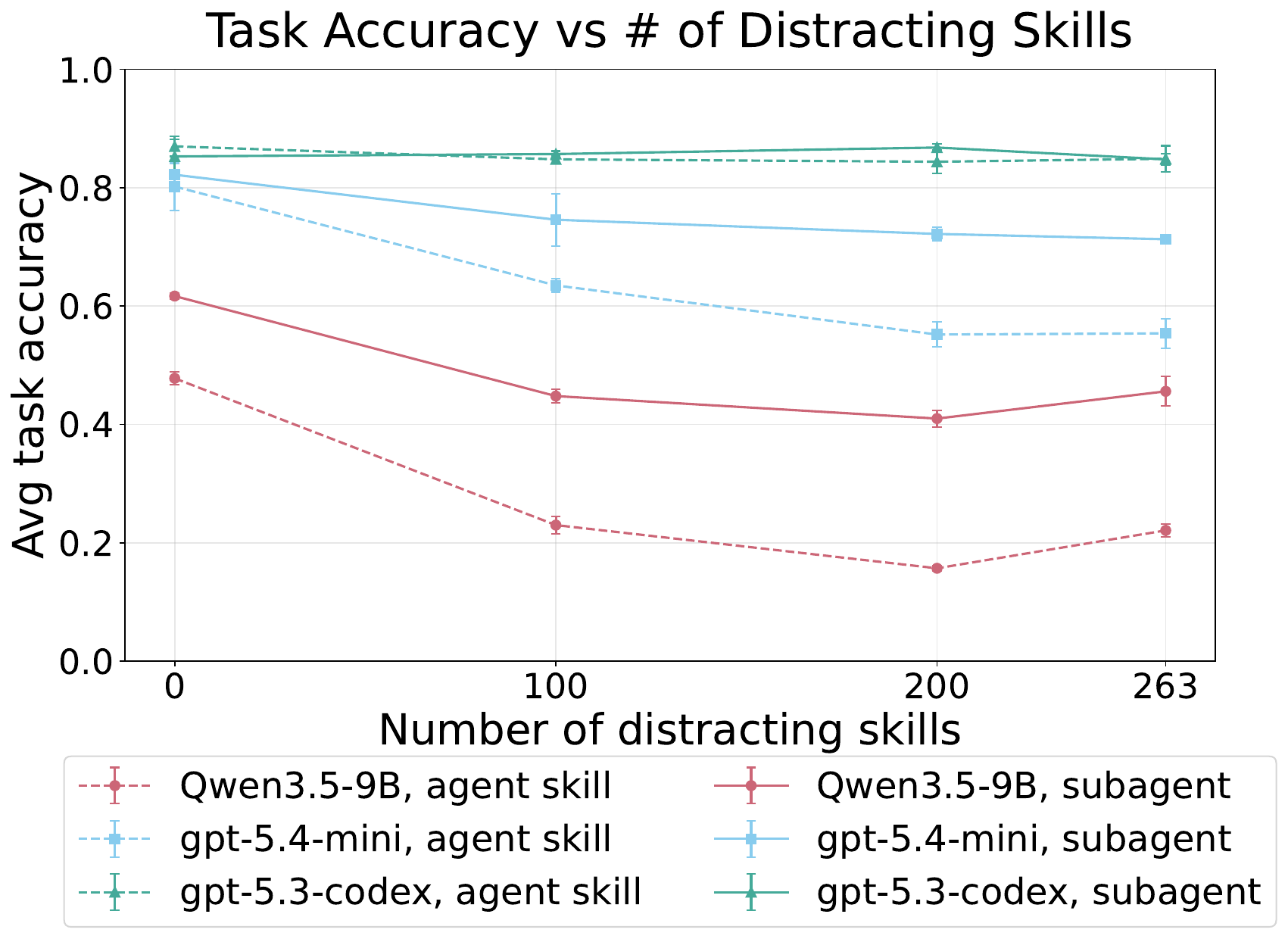}
\caption{Average task accuracy vs number of distracting skills for each base LLM model and execution mode. The skill packages used are from our synthesized set. Only three LLMs are displayed here for readability, see the full plot at \Cref{fig:scope_full}.}
\label{fig:scope}
\end{wrapfigure}

\paragraph{Scaling with increasing context information.}\label{sec:exp_scaling}
We next study how both execution strategies scale as input size grows and context pressure intensifies. 
To do so, we introduce distracting tools whose descriptions are appended to the agent context but are not required for solving the task. 
This simulates a realistic setting in which agents operate in environments containing many available tools. 
As the number of distracting tools increases, the initial context becomes substantially longer.

\Cref{fig:scope} (right) shows that subagent execution degrades more gracefully as the number of distracting tools increases. This confirms the benefit of peak context reduction discussed in \Cref{sec:reduce_peak}: keeping skill instructions out of an already-large context matters even more as that context grows.

\begin{figure*}[t]
\centering
\includegraphics[width=.47\linewidth]{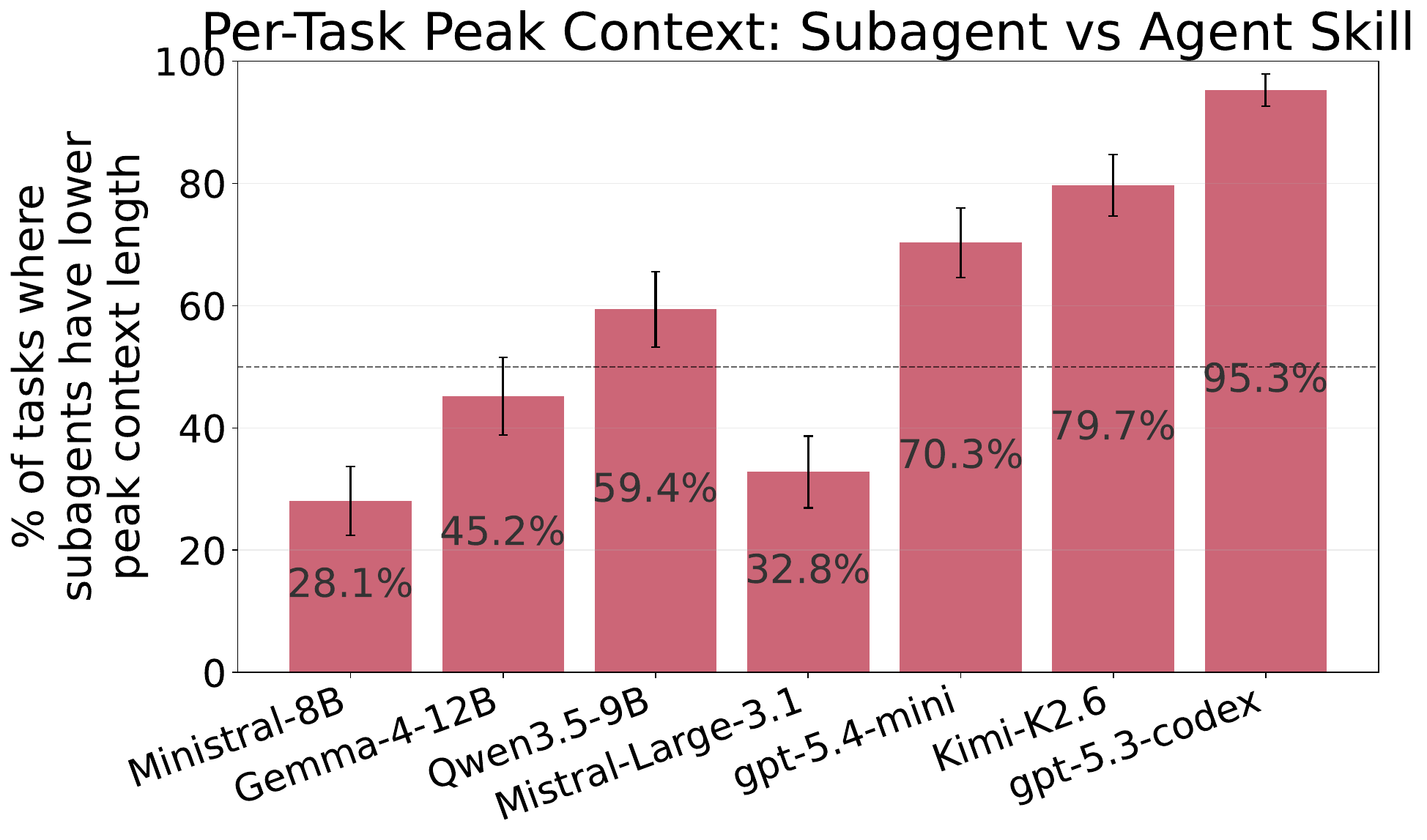}
\includegraphics[width=.52\linewidth]{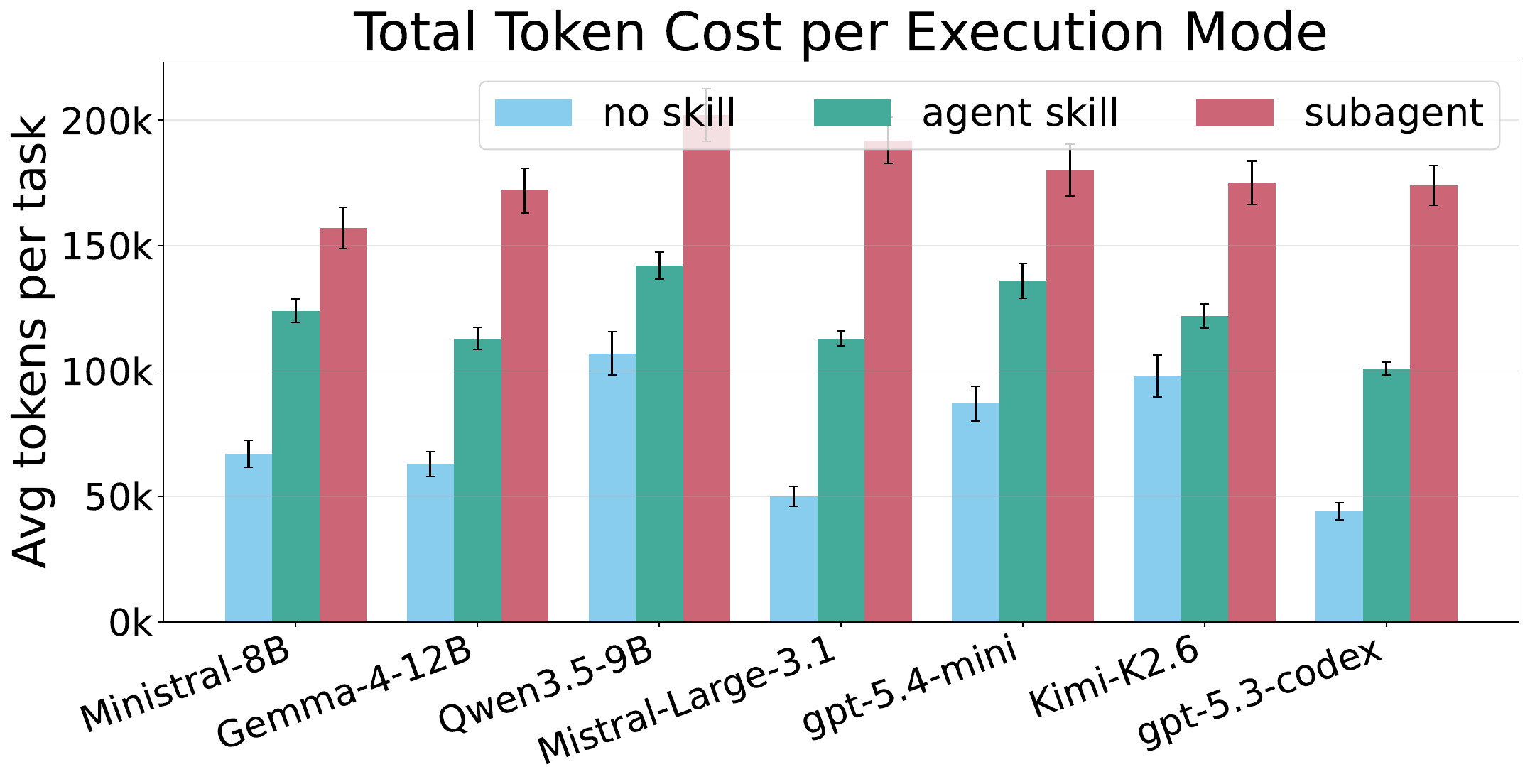}
\vspace{-1.5em}
\caption{Left: The percentage of tasks whose peak context length used is lower for subagent execution than agent skill execution.
Right: Total token used by each execution mode for each LLM model.
The skill packages used are from our synthesized set for both left and right figures.}
\label{fig:peak_and_cost}
\end{figure*}

\paragraph{Peak context length and total token cost.}
We now test our claim (\Cref{sec:reduce_peak}) that subagents trade communication overhead for reduced peak context.

\Cref{fig:peak_and_cost} (left) shows that for the stronger models, such as GPT 5.3 Codex and Kimi K2.6, subagents lower peak context length on over 80\% of tasks, as predicted.

For weaker models, the reduction is smaller and sometimes reverses.
This appears to reflect early termination under agent-skill execution: as the agent's reasoning degrades, the run ends without spending enough reasoning effort on the task, so its context stays short.
Subagent runs behave differently---each subagent works on a single coherent subtask and can spend the tokens needed to complete it.
Peak context length is therefore only comparable across models whose success rates are similar under both execution modes, which the stronger models satisfy and the weaker ones do not.

\Cref{fig:peak_and_cost} (right) shows the corresponding token cost. Subagent execution consumes substantially more tokens overall, as expected: an agent skill leaves all information in a single context, while each subagent must be supplied the context it needs, duplicating information the main agent already holds.





\section{Related Work}

\paragraph{Tool-Calling Language Model Agents.}

LLMs have become powerful agents when augmented with tools.
ReAct interleaves tool use with reasoning through prompting \cite{yao2023react}, while Toolformer, Gorilla, and ToolLLM instead train models to use tools, fine-tuning them to decide when to call a tool and to generate correct calls \cite{schick2023toolformer, patil2024gorilla, qin2024toolllm}. 
Closely related to our work is HuggingGPT, which uses other LLMs as tools, orchestrated by a central LLM controller \cite{shen2023hugginggpt}, similar to how subagents act as LLM-based tools in our setting; unlike our work, however, it does not study skill package execution.

While early tool calling benchmarks evaluated single-turn API routing \cite{li2023api, patil2025berkeley}, real-world use increasingly demands long, stateful interactions across many tools and applications \cite{yao2024tau, barres2025tau2, li2026skillsbench, li2026tool}.
Tasks in SkillsBench, for example, require many tool-calling turns, accumulate substantial information in context, and thus demand methods that scale with context size.

\paragraph{LLM Skill Learning.}

Skill learning for LLM agents originally focused on programmatic skills.
Voyager \cite{wang2023voyager} grows a library of executable code skills for a Minecraft agent.
LATM \cite{cai2024large}, CREATOR \cite{qian2023creator}, and CRAFT \cite{yuan2024craft} instead frame skill acquisition as tool creation, having an LLM author reusable code tools.
More recently, SkillCraft \cite{chen2026skillcraft} learns programmatic skills by composing a sequence of tool calls into skills and benchmarks skill reuse.

Following the wide adoption of skill packages as a skill format \cite{anthropic2025skills}, we have seen a surge of skill-package learning papers that construct or refine multi-file skill folders directly from agent experience \cite{alzubi2026evoskill, zhang2026coevoskills, ni2026trace2skill}. 
At their core, these methods mine execution trajectories or failures and use them to synthesize skill packages. 
These works all execute skill packages as agent skills.
Our work also uses skill packages but does not focus on skill-package learning; instead, we show that when skill packages are written or learned to follow a certain design, subagents can be used effectively, and they scale well with context size.

\nocite{xia2026skillrl}
\nocite{wang2026skillx}

\section{Conclusion and Future Work}

We studied how reusable skill packages should be executed in long-horizon agentic tasks. While agent skills execute by loading skill instructions into the main agent context, subagents execute skills in separate contexts and return only their outputs. We find that when context overload is an issue, subagent execution performs better and degrades more gracefully than agent skill execution as initial context scales.
By distributing computation across multiple context windows, subagents reduce the amount of information that any single context must process.

However, there are limitations to subagent execution. Our experiments suggest that subagents work well only when skill packages encode procedural knowledge with clear input-output contracts. Such skills can be executed largely independently of the main agent's context, making them suitable for encapsulation in a separate context. Existing curated skills do not always exhibit this property. Nevertheless, we show that procedural, contract-driven skills can be synthesized from successful trajectories, suggesting a practical pathway for learning skills that are suited to subagent execution.

More broadly, our results suggest that scaling agent systems requires reducing context load. Modularity helps control information flow by encapsulating task-specific knowledge behind clear interfaces.
From this perspective, subagent execution can be viewed as a mechanism for enforcing modularity.

Looking forward, two directions deserve further investigation. 
First, while subagent execution reduces peak context length, it incurs additional communication cost, raising the question of how the main agent and subagents should communicate efficiently---designing skills with minimal contracts is one concrete route, but the general problem remains open.
Second, we have only begun to explore the problem of skill library organization (\Cref{app:library}). 
Future work could investigate the notion of skill-library refactoring: how should reusable procedures be abstracted, factored, and organized as a skill library grows? 
In the same way that software engineering studies how large codebases should be structured, scalable agent systems may require principled approaches for organizing procedural knowledge hierarchically.

\bibliographystyle{unsrt}
{\small \bibliography{citations} }

\clearpage


\appendix

\section{Appendix}

\subsection{Organizing Libraries of Skills to Decrease Initial Context Load}\label{app:library}

We next ask whether hierarchically organizing a skill library can amplify the
benefits of context isolation. Recent agent systems increasingly lazy-load
MCP servers, exposing capabilities only when needed, which keeps the initial
context short and lowers peak context length in later turns. We test an
analogous strategy for skill libraries: rather than exposing every skill at
once, we organize skills into a hierarchy and expose only a subset at each
level.

\paragraph{Library organization.}
When a library contains many skills, exposing all of them to the agent at
once can be overwhelming for the agent. 
A hierarchical library instead exposes
only a small set of top-level nodes at first; selecting a node either
performs a task (if it is a leaf skill) or reveals a further set of
child nodes to choose from (if it is a \emph{routing} node). 
The agent
navigates the hierarchy step by step rather than seeing the whole library
up front. 
We compare four different library organizations, all with the leaf nodes being the well-interfaced procedural skills evaluated in \Cref{fig:main}:
\begin{itemize}
    \item \textbf{Flat library.} No hierarchy: all leaf skills are exposed
    to the agent simultaneously. 
    This is the baseline used throughout
    \Cref{fig:main}.
    \item \textbf{Hierarchical tree library.} 
    An LLM groups the leaf
    skills into a tree of routing nodes, where each routing node has
    exactly one parent, so there is a single path from the root to any
    leaf.
    \item \textbf{Hierarchical graph library.} The same LLM-built
    organization as the tree library, except a node may have multiple
    parents, so the structure is a DAG rather than a tree and a leaf can
    be reached via more than one path.
    \item \textbf{Two-level task-tree library.} A simpler
    baseline (no LLM involved): leaf skills are grouped by the task they
    belong to, giving a fixed two-level hierarchy of task node to leaf skills.
\end{itemize}

\paragraph{Execution modes.}
Independently of library structure, we vary how nodes are executed once
selected:
\begin{itemize}
    \item \textbf{Fully agent skill:} every node, routing or leaf, runs
    inline in the main agent's context.
    \item \textbf{Fully subagent:} every node runs as an isolated subagent
    call.
    \item \textbf{Hybrid:} routing (higher-level) nodes run as agent
    skills, while leaf skills run as subagents.
\end{itemize}

\begin{figure*}[t]
\centering
\includegraphics[width=\linewidth]{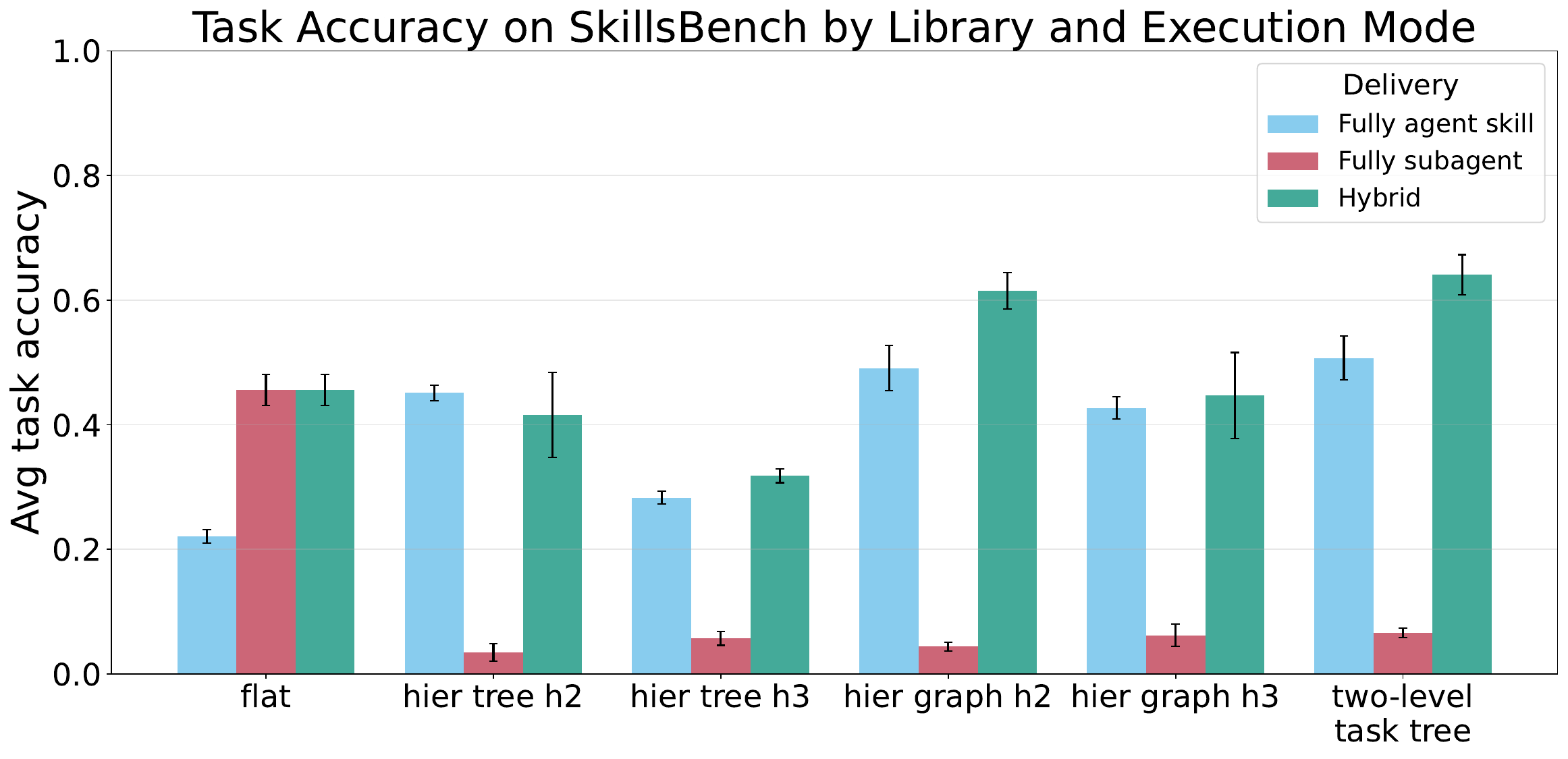}
\caption{Average task accuracy on SkillsBench for each library organization (flat, hierarchical tree, hierarchical graph, and two-level task tree) and execution mode (fully agent skill / fully subagent / hybrid) on Qwen3.5-9B. 
\emph{h2} and \emph{h3} means height=2 and height=3 respectively.}
\label{fig:library}
\end{figure*}

\paragraph{Results.}
\Cref{fig:library} shows that hierarchical organization gives additional
gains on top of our already well-interfaced procedural skill packages. The
task-tree and hierarchical-graph libraries perform best, and the LLM-built
graph outperforms the LLM-built tree, suggesting that allowing multiple
paths to a leaf skill helps. We did not observe further gains from deeper
hierarchies in our experiments, though we expect
benefits from deeper structure to emerge with libraries of thousands of
skills.

Across all four library types, the hybrid execution mode is consistently
best: routing nodes as agent skills, leaf skills as subagents. This mirrors
lazy-loading in MCP-based harnesses and supports our broader
\emph{effective subagent hypothesis}: routing nodes carry no procedural
knowledge or clear input-output contract, so they are better executed
inline as agent skills, while leaf skills -- which do have such contracts --
benefit from running as isolated subagents. We leave a systematic study of
deeper hierarchies and tree-vs-graph structure to future work.

\subsection{Additional results on SkillsBench}

Here, we include the average task accuracy vs number of distracting skills plot, for \Cref{sec:exp_scaling}, with more base models.
The interpretation of the results remains the same; we choose to only include only three LLMs in the main text for better readability.

\begin{figure*}[t]
\centering
\includegraphics[width=\linewidth]{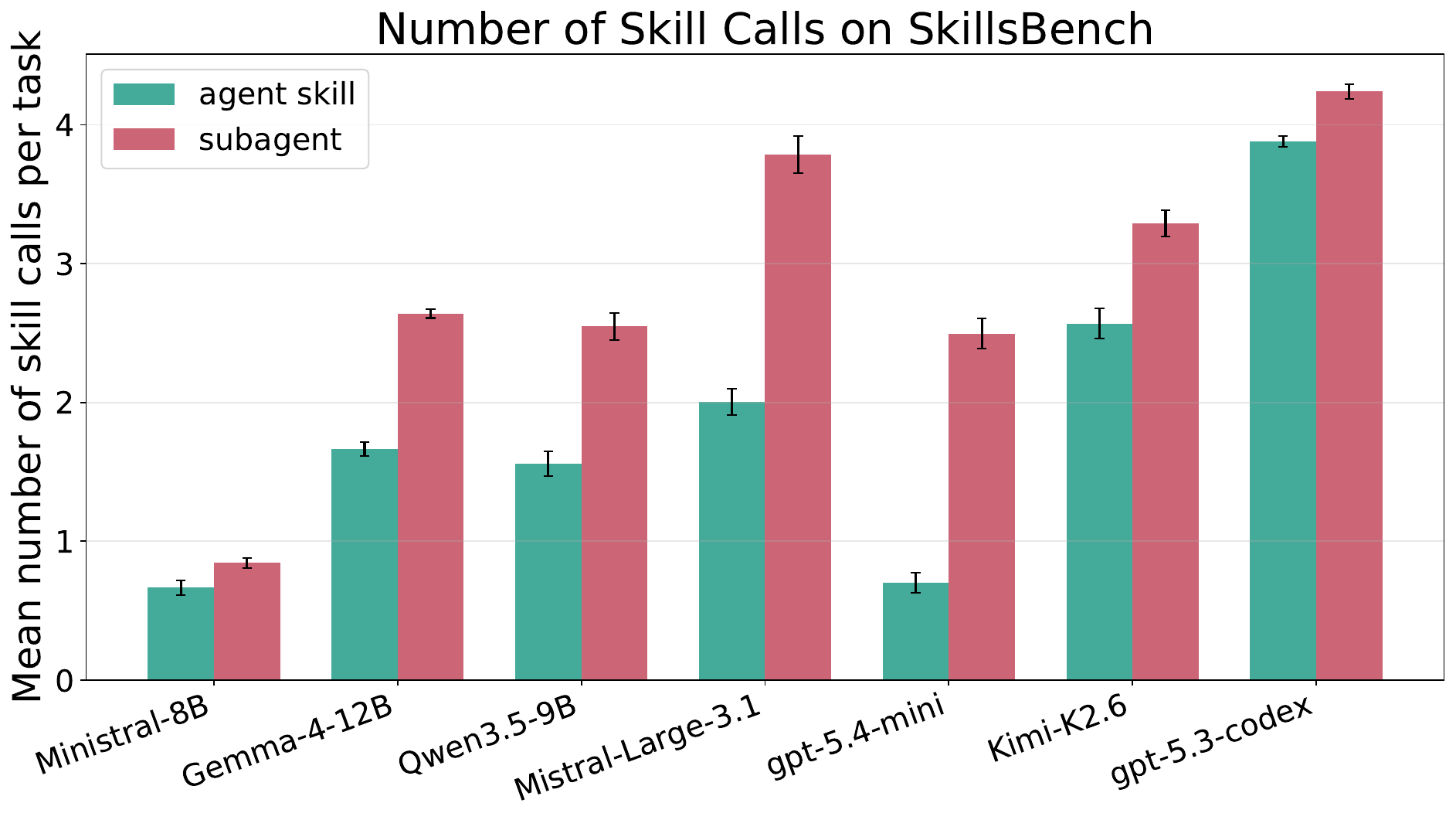}
\caption{Number of skill calls for each base LLM model. The skill packages used are
from our synthesized set.}
\label{fig:skill_invokation}
\end{figure*}

We also report number of skill calls per task at \Cref{fig:skill_invokation}.
We see that subagent execution leads to more skill calls per task across all base LLMs.
With agent-skill execution, we think that the LLMs start to underthink as their context grows \cite{liu2025llmigrate}, leading to lower skill calls on average.

\begin{figure*}[t]
\centering
\includegraphics[width=\linewidth]{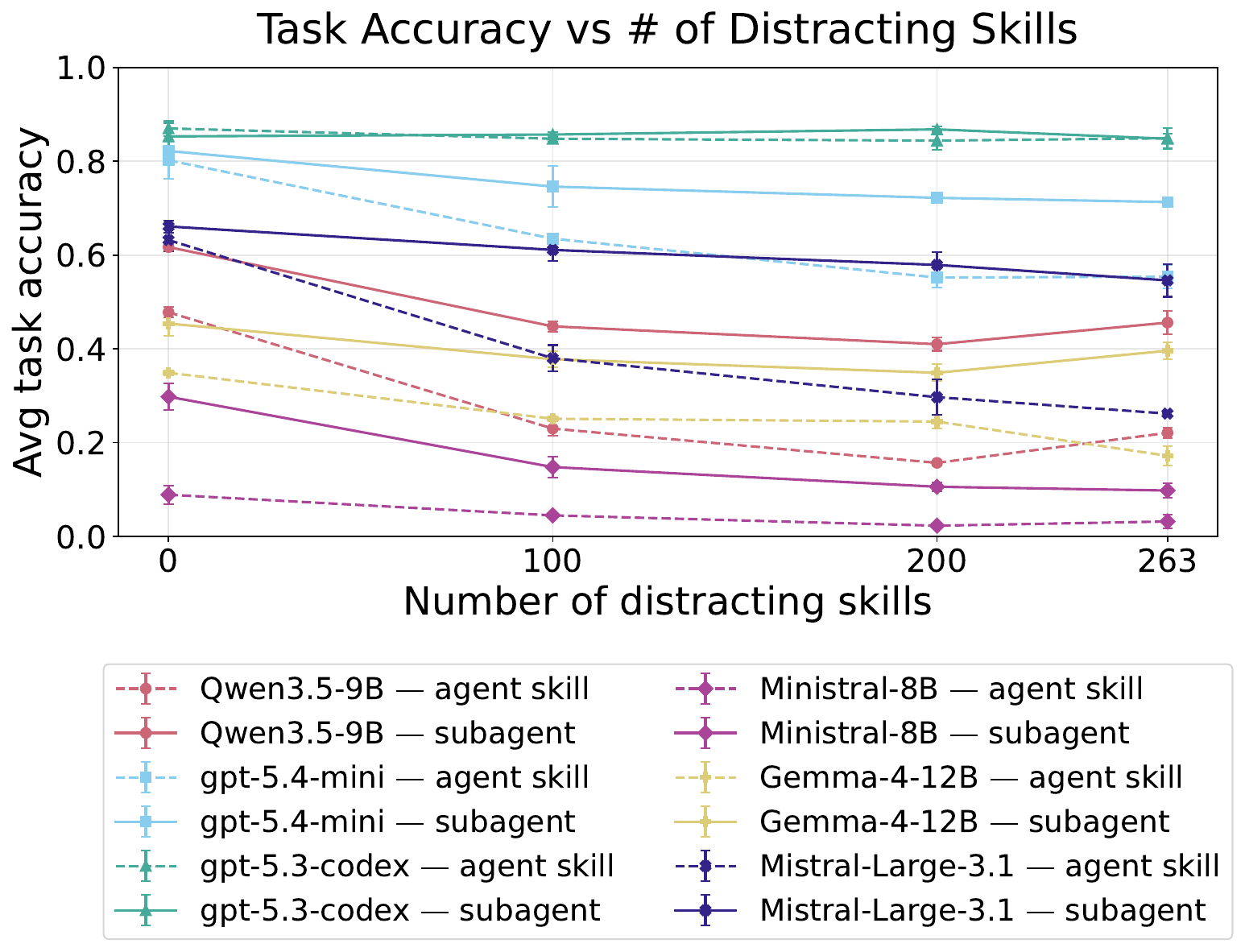}
\caption{Average task accuracy vs number of distracting skills for each base LLM model and execution mode. The skill packages used are from our synthesized set.}
\label{fig:scope_full}
\end{figure*}

\newtcolorbox{promptbox}[2][]{
  breakable,
  enhanced,
  colback=gray!4,
  colframe=black!55,
  colbacktitle=black!80,
  coltitle=white,
  fonttitle=\bfseries\small,
  boxrule=0.4pt,
  arc=1mm,
  left=4pt, right=4pt, top=3pt, bottom=3pt,
  title=#2,
  #1
}

\subsection{OpenHands Agent Implementation Details}
\label{app:openhands}

We use OpenHands~\cite{wang2025openhands} as one of the underlying agent frameworks in
our experiments.
When running each SkillsBench task, we spin up a docker container and run OpenHands agent inside the task's Docker container as the benchmark harness's registered
agent.
We route every LLM call through a proxy that we control, which lets us inject system prompts, apply sampling overrides, and log every request/response pair
uniformly. 

\paragraph{Prompt templates.}

We replace the Openhands' own vanilla system prompt with a much simpler prompt shown at \Cref{fig:oh-prompt-simple}.
When subagents get spawned, they are initialized with the prompt shown at \Cref{fig:oh-prompt-subagent}.

\begin{figure}[htbp]
\centering
\begin{promptbox}{Default system prompt (\texttt{simple})}
\begin{verbatim}
You are AI agents with access to tools. Please solve the
user-specified task as best as you can. Please always use
tools whenever they are relevant -- don't try to solve
subtasks on your own. Please try to use the <tool1>, <tool2>, ... 
when possible.
\end{verbatim}
\end{promptbox}
\caption{The system prompt used for our OpenHands agent. \emph{tool1}, \emph{tool2}, and so on list the name of the skills (agent skills / subagents) available.}
\label{fig:oh-prompt-simple}
\end{figure}

\begin{figure}[htbp]
\centering
\begin{promptbox}{Subagent task template}
\begin{verbatim}
You will be given a task and the knowledge on how to solve it

TASK:
{task}

INPUT INFORMATION:                 [only when input_info is non-empty]
{input_info}

KNOWLEDGE:
{skill_body}

Resources mentioned in KNOWLEDGE are available at: {skill_dir}

IMPORTANT: When you are done, return the output this skill
produces in your FINAL message. If the output is a file (or
files) you created or edited, state its exact path in your
final message; otherwise put the actual result text in your
final message. Do not end without reporting the output.
\end{verbatim}
\end{promptbox}
\caption{Initial prompt for a spawned subagent. \emph{skill\_body} is the content in a SKILL.md file. \emph{task} and \emph{input\_info} are given to the subagent by the main agent.}
\label{fig:oh-prompt-subagent}
\end{figure}

\paragraph{Other modifications to vanilla OpenHands SDK}

We also implement two modifications to the vanilla OpenHand SDK. First is we extend the subagent timeout. The OpenHands SDK hardcodes a 5-min timeout on every MCP tool call. Since our subagents are treated as MCP tools, we extend this timeout to two hours.
Second is we remove OpenHands' native skills. We also disable OpenHands' native auto-skill retrieval.


\begin{figure}[htbp]
\centering
\begin{promptbox}{Skill vulnerability-record-normalization description}
\begin{verbatim}
Convert scanner JSON into normalized CSV-ready vulnerability records 
using the task severity filter and deterministic CVSS/fixed-version 
fallbacks.
Expected input:
    - the scanner JSON report path from the scan step 
    - the required severity filter and CSV columns from the audit plan 
Output: 
    - a JSON list of normalized records with Package, Version, CVE_ID, 
    Severity, CVSS_Score, Fixed_Version, Title, and Url fields.
\end{verbatim}
\end{promptbox}
\caption{A description of a well-interfaced procedural skill, vulnerability-record-normalization}
\label{fig:skill1}
\end{figure}

\begin{figure}[htbp]
\centering
\begin{promptbox}{Skill offline-vulnerability-scan description}
\begin{verbatim}
Run a reproducible vulnerability scan for a dependency lockfile and save 
the full scanner result as JSON for later normalization. 
Expected input:
    - the audit plan containing the dependency lockfile path and desired
    JSON report path 
    - the scanner/cache strategy, including any local Trivy cache 
    directory if available 
Output: 
    - a JSON vulnerability report saved on disk plus a summary of total 
    findings and counts by severity.
\end{verbatim}
\end{promptbox}
\caption{A description of a well-interfaced procedural skill, offline-vulnerability-scan}
\label{fig:skill2}
\end{figure}

\subsection{Well-interfaced Procedural Skill Authoring details}\label{app:skill_learning}

As mentioned in \Cref{sec:exp}, we mainly use Copilot CLI, powered by Claude Opus 5, to come up with our well-interfaced procedural skill packages.
Initially, we ask Copilot CLI to come up with a set of 3-5 of such skill packages for a single task, called edit-pdf. 
Then, we manually intervene to ensure the the synthesized skills do have an "expected input" and "output" description, as part of their skill description, and the skill instructions faithfully try to transform the input to output.
Then, we ask Copilot CLI to spawn subagents to do write a set of 3-5 well-interfaced procedural skill packages for each of the task with successful trajectories, using the skill set from the initial task, edit-pdf, as an example.
Copilot CLI uses GPT-5.5 to power the subagents.

We show an example output skill descriptions from this process at \Cref{fig:skill1,fig:skill2}.

\subsection{Computational Resources}
\label{app:oh-compute}

Experiments were run on NVIDIA A100 GPU nodes.
Open-weight models
(the Qwen3.5 family at 2B/4B/9B parameters, Ministral-3-8B, and
Gemma-4-12B-it) were served locally with vLLM directly on an A100 node's GPU.
Closed-weight or large models(the GPT-5.x family, Mistral-Large-3, Kimi-K2.6) were
accessed through a hosting API and requires no GPU itself.



\end{document}